\documentclass[letterpaper, 10 pt, conference]{ieeeconf}  % Comment this line out if you need a4paper

\IEEEoverridecommandlockouts                              % This command is only needed if
\usepackage{authblk}
\usepackage{graphics} % for pdf, bitmapped graphics files
\usepackage{epsfig} % for postscript graphics files
\usepackage{times} % assumes new font selection scheme installed
\usepackage{amsmath} % assumes amsmath package installed
\usepackage{amssymb}  % assumes amsmath package installed
\usepackage{breqn}
\usepackage{xcolor} % added by yuansi for TODO, to be deleted in final version.

\title{\LARGE \bf
Self-calibrating Neural Networks for Dimensionality Reduction
}

\author[1,2]{Yuansi Chen\thanks{yuansi.chen@berkeley.edu}}
\author[2]{Cengiz Pehlevan\thanks{cpehlevan@simonsfoundation.org}}
\author[2,3]{Dmitri B. Chklovskii \thanks{mitya@simonsfoundation.org}}
\affil[1]{Statistics Department of University, California, Berkeley, CA}
\affil[2]{Center for Computational Biology, Flatiron Institute, New York, NY}
\affil[3]{NYU Medical School, NYU, New York, NY}

\begin{document}

\maketitle
\thispagestyle{empty}
\pagestyle{empty}

%%%%%%%%%%%%%%%%%%%%%%%%%%%%%%%%%%%%%%%%%%%%%%%%%%%%%%%%%%%%%%%%%%%%%%%%%%%%%%%%
\begin{abstract}

Recently, a novel family of biologically plausible online algorithms for reducing the dimensionality of streaming data has been derived from the similarity matching principle. In these algorithms, the number of output dimensions can be determined adaptively by thresholding the singular values of the input data matrix. However, setting such threshold requires knowing the magnitude of the desired singular values in advance. Here we propose online algorithms where the threshold is self-calibrating based on the singular values computed from the existing observations. To derive these algorithms from the similarity matching cost function we propose novel regularizers. As before, these online algorithms can be implemented by Hebbian/anti-Hebbian neural networks in which the learning rule depends on the chosen regularizer. We demonstrate both mathematically and via simulation the effectiveness of these online algorithms in various settings.
\end{abstract}

%%%%%%%%%%%%%%%%%%%%%%%%%%%%%%%%%%%%%%%%%%%%%%%%%%%%%%%%%%%%%%%%%%%%%%%%%%%%%%%%

% 1. Introduction
\section{Introduction}

Dimensionality reduction plays an important role in both artificial and natural signal processing systems. In man-made data analysis pipelines it denoises the input, simplifies further processing and identifies important features. Dimensionality reduction algorithms have been developed for both the offline setting where the whole dataset is available to the algorithm from the outset and the online setting where data are streamed one sample at a time \cite{van2009dimensionality,yan2006effective}. In the brain, online dimensionality reduction takes place, for example, in early processing of streamed sensory inputs as evidenced by a high ratio of input to output nerve fiber counts \cite{hyvarinen2009natural}. Therefore, dimensionality reduction algorithms may help model neuronal circuits and observations from neuroscience may inspire the future development of artificial signal processing systems.

Recently, a novel principled approach to online dimensionality reduction in neuronal circuits has been developed \cite{pehlevan2015hebbian,hu2014hebbian, pehlevan2014hebbian, pehlevan2015normative}. This approach is based on the principle of similarity matching: the similarity of outputs must match the similarity of inputs under certain constraints. Mathematically, pairwise similarity is quantified by the inner products of the data vectors and matching is enforced by the  classical multidimensional scaling (CMDS) cost function. Dimensionality is reduced by constraining the number of output degrees of freedom, either explicitly or by adding a regularization term.

To formulate similarity matching mathematically, we represent each centered input data sample received at time $t$ by a column vector $\mathbf{x}_t \in \mathbb{R}^n$ and the corresponding output by a column vector, $\mathbf{y}_t \in \mathbb{R}^k$. We concatenate the input vectors into an $n\times T$ input data matrix $\mathbf{X} = [\mathbf{x}_1, ..., \mathbf{x}_T]$ and the output vectors into an $k\times T$ output data matrix $\mathbf{Y} = [\mathbf{y}_1, ..., \mathbf{y}_T]$. Then we match all pairwise similarities by minimizing the summed squared differences between all pairwise similarities (known as the CMDS cost function, \cite{carroll1972idioscal,mardia1980multivariate,cox2000multidimensional}):
\begin{equation}\label{eq_CMDS_cost}
\min_{\mathbf{Y}} \left\| \mathbf{X}^\top \mathbf{X} - \mathbf{Y}^\top\mathbf{Y} \right\|_F^2
\end{equation}
To avoid the trivial solution, $\mathbf{Y} = \mathbf{X}$, we restrict the number of degrees of freedom in the output by setting $k<n$. Then the solution to the minimization problem (\ref{eq_CMDS_cost}) is the projection of the input data on the $k$-dimensional principal subspace of $\mathbf{X}^\top\mathbf{X}$ \cite{mardia1980multivariate,cox2000multidimensional}, i.e. the subspace spanned by the eigenvectors corresponding to the top $k$ eigenvalues of the input similarity matrix $\mathbf{X}^\top\mathbf{X}$.

In  \cite{pehlevan2015hebbian,hu2014hebbian, pehlevan2014hebbian}, the similarity matching objective (\ref{eq_CMDS_cost}) was optimized in the online setting, where input data vectors arrive sequentially, one at a time, and the corresponding output is computed prior to the arrival of the next input. Remarkably, the derived algorithm can be implemented by a single-layer neural network (Figure~\ref{fig:network}, left), where the components of input (output) data vectors are represented by the activity of input (output) neurons at the corresponding time. The algorithm proceeds in alternating steps where the neuronal dynamics computes the output $\mathbf{y}_T$ and the synaptic weights are updated according to local learning rules, meaning that a weight update of each synapse depends on the activity of only its pre- and postsynaptic neurons. The family of similarity matching neural networks  \cite{pehlevan2015hebbian,hu2014hebbian, pehlevan2014hebbian, pehlevan2015normative} is unique among dimensionality reduction networks in combining biological plausibility with the derivation from a principled cost function.

In real-world signal-processing applications and neuroscience context the desired number of output dimensions is often unknown to the algorithm a priori and varies with time because of input non-stationarity. Because the number of output dimensions in the neural circuit solution of (1) is the number of output neurons it cannot be adjusted quickly.

To circumvent this problem, we proposed to penalize the rank of the output by adding a regularizer $R_{T\mathbf{Y}}=\alpha T{\mathrm Tr}(\mathbf{Y}^\top \mathbf{Y})= \alpha T\left \| \mathbf{Y}^\top \mathbf{Y} \right\|_*$ where $\left \| .\right \|_*$  is a nuclear norm of a matrix known to be a convex relaxation of matrix rank. From the regularized cost function we derived adaptive algorithms \cite{pehlevan2015normative} in which the number of output dimensions is given by the number of input singular values exceeding a threshold that depends on the parameter $\alpha$. Because singular values scale with the number of time steps $T$ the threshold scales with $T$.  %Such formulation is appropriate when the noise is white, \todoyuansi{mention Donoho. Also mention Tao's sparsity inducing regularizer}.
However, choosing a regularization parameter is hard because it requires knowing the exact scale of input singular values in advance. Furthermore, a scale-dependent threshold is not adaptive to non-stationary inputs with changing singular values.

In this paper, we introduce {\it self-calibrating} regularizers for the cost function (\ref{eq_CMDS_cost}) which do not depend on time explicitly and are designed to automatically adjust to the variation in singular values of the input. Specifically, we propose $R_{\mathbf{XY}}= \alpha {\mathrm Tr}(\mathbf{X}^\top \mathbf{X}){\mathrm Tr}(\mathbf{Y}^\top \mathbf{Y})$ and $R_{\mathbf{YY}}= \alpha ({\mathrm Tr}(\mathbf{Y}^\top \mathbf{Y}))^2$. We solve the cost function with these regularizers in both offline (Section II) and online (Section IV) settings. These two online algorithms also map onto a single-layer neuronal network but, importantly, with different learning rules. In Section III, we mathemathically illustrate the difference among the three regularizers in a very simple data generation scenario. In Section V, we propose the corresponding algorithms for the non-stationary input scenario by introducing discounting or forgetting.

%%%%%%%%%%%%%%%%%%%%%%%%%%%%%%%%%%%%%%%%%%%%%%%%%%%%%%%%%%%%%%%%%%%%%%%%%%%%%%%%

% 2. Offline adaptive dimension reduction

\section{Adaptive dimension reduction in the offline setting}
In this section, we first summarize previous derivation of adaptive dimensionality reduction from cost function (\ref{eq_CMDS_cost}) with a scale-dependent regularizer, $R_{T{\mathbf{Y}}}$, \cite{pehlevan2015normative} and discuss its potential shortcomings. Then we introduce two self-calibrating adaptive dimension reduction methods, involving solving cost function (\ref{eq_CMDS_cost}) offline with two new regularizers $R_{\mathbf{XY}}$ and $R_{\mathbf{YY}}$.

\subsection{Scale-dependent regularizer,  $R_{T{\mathbf{Y}}}$}
% fixed \todoyuansi{CP:I don't like the name scale-dependent regularizer. How about scale-dependent regularizer?}

In order to adaptively choose the output dimension, \cite{pehlevan2015normative} proposed to modify the objective function (\ref{eq_CMDS_cost}) by adding a scale-dependent regularizer:
\begin{equation}\label{eq_offline_TY}
\min_{\mathbf{Y}} \left\| \mathbf{X}^\top\mathbf{X} - \mathbf{Y}^\top\mathbf{Y} \right\|_F^2 + 2\alpha T {\mathrm Tr} ( \mathbf{Y}^\top\mathbf{Y} ),
\end{equation}
with $\alpha \geq 0$.
Such a regularizer corresponds to the trace norm of $\mathbf{Y}^\top\mathbf{Y}$ which is a convex relaxation of rank. Its impact on the solution can be better understood by rewriting the cost as a full square which has the same $\mathbf{Y}$-dependent terms:
\begin{equation}\label{eq_offline_TY_variant}
\min_{\mathbf{Y}} \left\| \mathbf{X}^\top\mathbf{X} - \mathbf{Y}^\top\mathbf{Y} - \alpha T \mathbf{I}_T\right\|_F^2,
\end{equation}
where $\mathbf{I}_T$ is a $T\times T$ identity matrix.

The optimal output matrix $\mathbf{Y}$ is a projection of the input data $\mathbf{X}$ onto its principal subspace \cite{pehlevan2015normative}, with soft-thresholding on the input singular values. Indeed, suppose the eigen-decomposition of $\mathbf{X}^\top\mathbf{X}$ is $\mathbf{X}^\top\mathbf{X} = \mathbf{V}^X \mathbf{\Lambda}^X {\mathbf{V}^X}^\top$, where $\mathbf{\Lambda}^X = \mathrm{diag}(\lambda_1^X, ..., \lambda_T^X)$ with $\lambda_1^X \geq ... \geq \lambda_T^X \geq 0$ are ordered eigenvalues of $\mathbf{X}^\top{\bf X}$.  Then the solution to the offline problem (\ref{eq_offline_TY}) is
\begin{equation}\label{eq_sol_offline_TY}
\mathbf{Y} = \mathbf{U}_k \mathrm{ST}_k(\mathbf{\Lambda}^X, \alpha T)^{1/2} {\mathbf{V}_k^X}^\top,
\end{equation}
where
\[
\mathrm{ST}_k({\mathbf{\Lambda}}^X,\alpha T) = {\rm diag} \left(\mathrm{ST}\left(\lambda^X_{1},\alpha T\right),\ldots,\mathrm{ST}\left(\lambda^X_{k},\alpha T \right)\right),
\]
$\mathrm{ST}$ is the soft-thresholding function, $\mathrm{ST}(a,b) = \max(a-b,0)$.  $\mathbf{V}_k^X$ consists of the columns of $\mathbf{V}^X$ corresponding to the top $k$ eigenvalues, i.e. $\mathbf{V}_k^X = [\mathbf{v}_1^X, ..., \mathbf{v}_k^X]$ and $\mathbf{U}_k$ is any $k \times k$ orthogonal matrix, i.e. $\mathbf{U}_k \in \mathcal{O}(k)$.

Equation (4) shows that the regularization coefficient $\alpha$ sets the threshold on the eigenvalues of input covariance. Input modes with eigenvalues above $\alpha T$ are included in the output, albeit with a eigenvalue shrunk by $\alpha T$. Modes below $\alpha T$ are rejected by setting corresponding output singular values to zero. The scaling of regularization coefficient with time, $T$, ensures that the threshold occupies the same relative position in the spectrum of eigenvalues, which grow linearly with time for a stationary signal.

The algorithm can separate signal from the noise if the signal eigenvalues are greater than the noise eigenvalues and $\alpha$ is set in between. However, setting such value of $\alpha$ requires knowing the variance of the input signal and noise. In the offline setting, $\alpha$ can be computed from the data. However, if $\alpha$ has to be chosen a priori, e.g. in the online setting, choosing the value that suits various inputs with different signal variance may be difficult. In particular, when the noise variance of one possible input exceeds the signal variance of another input, a universal value of $\alpha$ does not exist. Is it possible to regularize the problem so that the regularization coefficient is chosen only once and applies universally to inputs of arbitrary variance?

\subsection{Input-output regularizer, $R_{\mathbf{XY}}$}
A regularizer that applies a relative, rather than absolute, threshold to input singular values would be able to deal with various input setting. Rather than using the threshold depending on time $T$ explicitly, we set the threshold value proportional to the sum of eigenvalues of $\mathbf{X}^\top\mathbf{X}$. Formally, this leads to the following optimization problem with input-output regularizer $R_{\mathbf{XY}}$:
\begin{equation}\label{eq_offline_XY}
\min_{\mathbf{Y}} \left\| \mathbf{X}^\top\mathbf{X} - \mathbf{Y}^\top\mathbf{Y}\right\|_F^2+ 2\alpha \mathrm{Tr}(\mathbf{X}^\top\mathbf{X})\mathrm{Tr}(\mathbf{Y}^\top\mathbf{Y})
\end{equation}
which is equivalent to:
\begin{equation}\label{eq_offline_XY_variant}
\min_{\mathbf{Y}} \left\| \mathbf{X}^\top\mathbf{X} - \mathbf{Y}^\top\mathbf{Y} - \alpha \mathrm{Tr}(\mathbf{X}^\top\mathbf{X}) \mathbf{I}_T\right\|_F^2
\end{equation}
In the offline setting, the change relative to previous method is minor because we can always have a good choice of $\alpha$ to make the former formulation similar to the later after observing the whole $\mathbf{X}$. As a consequence, the offline solution of Eq. (\ref{eq_offline_XY}) is very similar to that of the previous method. $\mathbf{Y}$ is a projection of the input data onto its principal subspace with a different eigenvalue cutoff based on the input eigenvalue sum. In turn, the coefficient, $\alpha$, sets the threshold relative to the sum of eigenvalues of $\mathrm{Tr}(\mathbf{X}^\top\mathbf{X})$:
%fixed \todoyuansi{CP: Mean or total? If we want mean, we should change $\alpha\longrightarrow\alpha/n$ in the cost function. This carries on in the rest of the paper}:
\begin{equation}\label{eq_sol_offline_XY}
\mathbf{Y} = \mathbf{U}_k \mathrm{ST}_k(\mathbf{\Lambda}^X, \alpha \mathrm{Tr}(\mathbf{X}^\top\mathbf{X}) )^{1/2} {\mathbf{V}_k^X}^\top.
\end{equation}
Even though this solution looks very similar to the scale-dependent adaptive dimension reduction's offline solution (\ref{eq_sol_offline_TY}), as we will see in Section IV, the fact that the thresholding depends on the input singular values will allow corresponding online algorithms to calibrate to various input statistics.

\subsection{Squared-output regularizer, $R_{\mathbf{YY}}$}
An alternative way to apply a relative threshold to input singular values is to deploy a regularization proportional to the sum of eigenvalues of $\mathbf{Y}^T\mathbf{Y}$. When doing dimension reduction, the sum of eigenvalues of $\mathbf{Y}^T\mathbf{Y}$ is reflective of the sum of top eigenvalues of $\mathbf{X}^T\mathbf{X}$. This reasoning leads us to the following optimization problem with squared-output regularizer $R_{\mathbf{YY}}$:
\begin{equation}\label{eq_offline_XY}
\min_{\mathbf{Y}} \left\| \mathbf{X}^\top\mathbf{X} - \mathbf{Y}^\top\mathbf{Y} \right\|_F^2 +  \alpha [\mathrm{Tr}(\mathbf{Y}^\top\mathbf{Y})]^2.
\end{equation}

This optimization problem is not as simple as in the previous case. Yet, the optimal output $\mathbf{Y}$ is still a projection of input data but with an adaptively thresholded singular values of $\mathbf{X}$:
\begin{align}\label{eq_sol_offline_YY}
\mathbf{Y} = \mathbf{U}_k {(\mathbf{D}_k^Y)}^{1/2} {\mathbf{V}_k^X}^\top,
\end{align}
where $\mathbf{D}_k^Y = \mathrm{diag}(\lambda_1^Y, ..., \lambda_p^Y, 0, ..., 0)$ and
\begin{align}
\mathbf{D}_p^Y = (\mathbf{I}_p - \frac{\alpha}{1+\alpha p} \mathbf{1}_p\mathbf{1}_p^\top) \mathbf{\Lambda}_p^X.
\end{align}
The interger $p$ decides how many singular values of the input $\mathbf{X}$ are cut off. It is chosen as the largest integer in $\{1, ..., k\}$ such that all diagonal elements of $\mathbf{D}_p^Y$ are nonnegative.

More details about the derivation of this closed-form offline solution can be found in Appendix A. Intuitively, the amount of shrinkage still depends approximately on the sum of input eigenvalues but the sum is computed only on the top $p$ eigenvalues. Similar to the input-output regularizer, the regularization coefficient $\alpha$ sets the threshold relatively to the input statistics.

%%%%%%%%%%%%%%%%%%%%%%%%%%%%%%%%%%%%%%%%%%%%%%%%%%%%%%%%%%%%%%%%%%%%%%%%%%%%%%%%

% 3. Comparing of the three regularizer on simple cases
\section{Three methods on a simple case: two sets of degenerate eigenvalues with a gap}

To gain intuition about the three similarity matching algorithms, we compare their offline solutions for the input covariance with only two sets of degenerate eigenvalues. Suppose that the eigenvalues of the normalized input similarity matrix $\frac 1T \mathbf{X}^T\mathbf{X}$ are
\[
(\underbrace{a,...,a}_{n_1}, \underbrace{b,...,b}_{n_2})
\] with $a>b$ and $n_1+n_2 = n$. This kind of scenario models a situation when signal and noise eigenvalues are separated by a gap.

We ask when the output similarity matrix keeps track of the $n_1$ signal eigenmodes but rejects the $n_2$ noise eigenmodes. We first derive, for each method, the range of $\alpha$'s achieving this goal.
\begin{enumerate}
\item For the scale-dependent regularizer $R_{T\mathbf{Y}}$, according to Eq. (\ref{eq_sol_offline_TY}), it is sufficient and necessary to choose the regularization coefficient $\alpha$ between $a$ and $b$. Note that this regularization coefficient $\alpha$, unlike the following two, depends on the absolute scale of the noise level $b$.
\item For the input-output regularizer $R_{\mathbf{XY}}$, $\text{Tr}(\frac{1}{T}\mathbf{X}^\top\mathbf{X}) = n_1 a + n_2 b$. According to Eq. (\ref{eq_sol_offline_XY}), it is sufficient and necessary to choose $\alpha$ such that
\begin{align}
\frac{a}{n_1 a + n_2 b} \geq \alpha \geq \frac{b}{n_1 a + n_2 b}.
\end{align}
\item For the squared-output regularizer $R_{\mathbf{YY}}$, we are aiming at choosing $p = n_1$. According to Eq. (\ref{eq_sol_offline_YY}), it is sufficient and necessary to choose $\alpha$ such that
the $n_1+1$st output similarity matrix's eigenvalue is non-positive for $p = n_1 + 1$.
This results in
\begin{align}
\alpha \geq \frac{b}{(a-b)n_1}
\end{align}
\end{enumerate}
Table \ref{tab:summary_reg} summarizes the different ranges of regularization coefficients $\alpha$ with which three methods can keep track of the first $n_1$ eigenmodes and the resulting output similarity matrix's top eigenvalue.

\begin{table*}[t]
\centering
 \caption{Summary of regularization coefficient $\alpha$ choice}
 \begin{tabular}{|l | c | c|}
 \hline
Regularizer & Choice of $\alpha$ & Output top eigenvalue \\
 \hline
1. Scale-dependent Regularizer, $R_{\mathbf{TY}}$& $a \geq \alpha \geq b$ & $a - \alpha$  \\
  \hline
2. Input-output Regularizer, $R_{\mathbf{XY}}$ & $\frac{a}{n_1 a + n_2 b} \geq \alpha \geq \frac{b}{n_1 a + n_2 b}$ & $a  - \alpha (n_1 a + n_2 b)$  \\
  \hline
3. Squared-output Regularizer, $R_{\mathbf{YY}}$ & $\alpha \geq \frac{b}{(a-b)n_1}$ & $a (\frac{1}{1+\alpha n_1})$  \\
 \hline
 \end{tabular}
 \label{tab:summary_reg}
\end{table*}

\begin{figure}[ht]
\centering
\includegraphics[width=0.8\columnwidth]{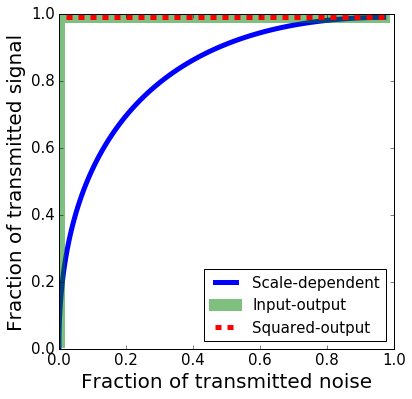}
\caption{Fraction of $a$ and $b$ pairs for which all signal is transmitted vs. fraction of $a$ and $b$ paris for which all noise is transmitted. Each curve is computed for $\alpha$ varying from $0$ to $\infty$ for each of the three regularizers. }
\label{fig:signal2noise}
\end{figure}

To illustrate the difference between the known scale-dependent regularizer and the two newly proposed regularizers we compute for what fraction of various pairs of $a$ and $b$ (see Appendix) each algorithm achieves the goal for values of $\alpha$ from $0$ to $\infty$. Fig.~\ref{fig:signal2noise} shows the fraction of $a$ and $b$ pairs for which the signal is transmitted vs. the fraction for which the noise is transmitted as $\alpha$ varies along each curve. The curve corresponding to the scale-dependent regularizer does not reach the point $(0, 1)$ in Fig.~\ref{fig:signal2noise} indicating that no value of $\alpha$ achieves the goal for all pairs of $a$ and $b$. Yet, the input-output and the squared-output regularizers pass through the point $(0,1)$ indicating that universal values of $\alpha$ exist for which these algorithms transmit all signal while discarding all noise.

\begin{figure}[ht]
\includegraphics[width=0.8\columnwidth]{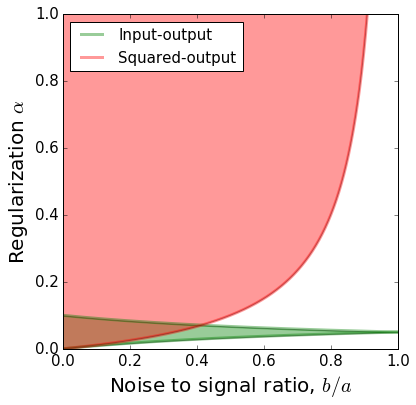}
\caption{Colored regions indicate the range of the regularization coefficient $\alpha$ for which all signal and no noise is transmitted. }
\label{fig:phase_diag}
\end{figure}

Nex, to illustrate the difference between the input-output and the squared-output regularizer we plot a phase diagram (Fig.~\ref{fig:phase_diag}) illustrating the range of parameters where each algorithm transmits all signal and rejects all noise. Specifically, Fig.~\ref{fig:phase_diag} shows the range of regularization coefficient $\alpha$ for different noise-to-signal ratio $b/a$ of the input.  The range of $\alpha$ for which the squared-output regularizer achieves the goal is much larger than that for the input-output regularizer indicating its robustness.

%%%%%%%%%%%%%%%%%%%%%%%%%%%%%%%%%%%%%%%%%%%%%%%%%%%%%%%%%%%%%%%%%%%%%%%%%%%%%%%%

\begin{figure*}
\centering
\includegraphics[width=0.9\textwidth]{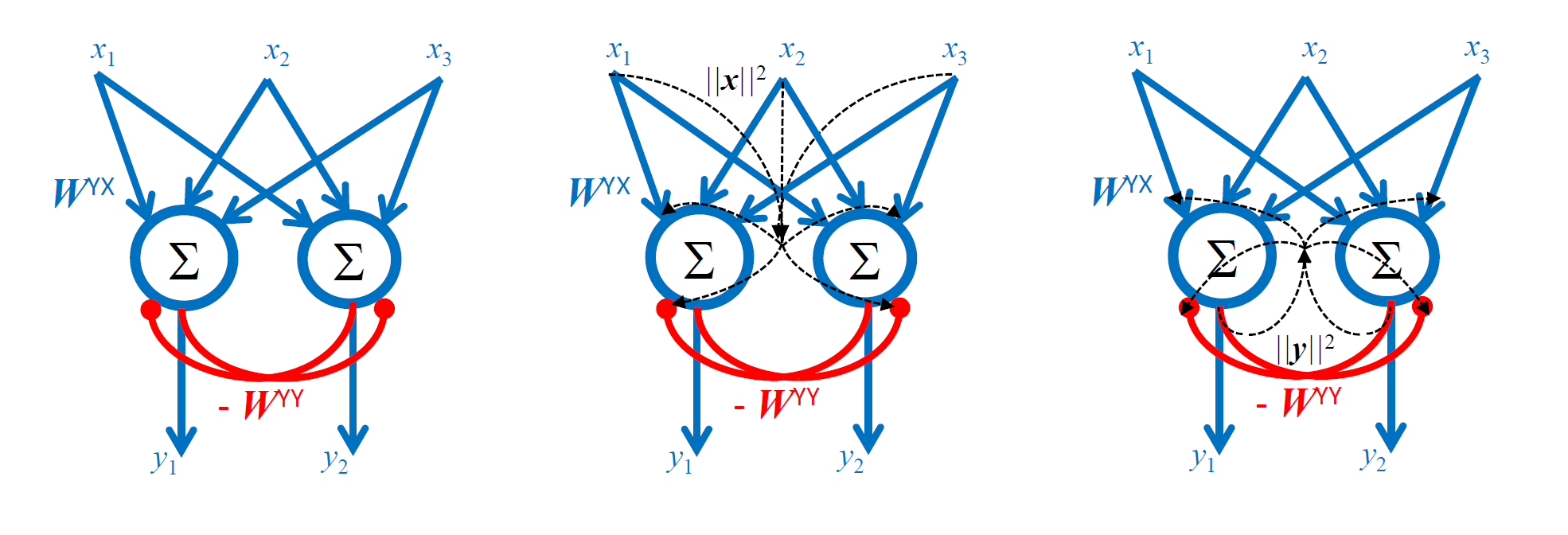}
\caption{Left: Hebbian/anti-Hebbian neural network implementing the online soft-thresholding algorithm with scale-dependent regularizer. Middle: same network with input-output regularizer. Right: same network with squared-output regularizer.}
\label{fig:network}
\end{figure*}

% 4. Online adaptive dimension reduction
\section{Online adaptive dimensionality reduction with Hebbian/anti-Hebbian neural networks}
In this section, we first formulate online versions of the dimensionality reduction optimization problems presented in Section II. Then we derive corresponding online algorithms and map them onto the dynamics of neural networks with biologically plausible local learning rules. Our derivations follow \cite{pehlevan2015normative}. At time $T$, the algorithm minimizes the cost depending on the previous inputs and outputs up to time $T$ with respect to $\mathbf{y}_T$, while keeping the previous $\mathbf{y}_t$ fixed.
\begin{equation}\label{eq_online_CMDS_cost}
\min_{\mathbf{y}_T} \left\| \mathbf{X}^\top \mathbf{X} - \mathbf{Y}^\top\mathbf{Y} \right\|_F^2
\end{equation}
In the unregularized formulation, the output dimensionality is determined by the column dimensition of $\mathbf{Y}$, $k$. Online adaptive dimension reduction methods adaptively choose the output dimensionality based on the trade-off of the CMDS cost with regularizers.

\subsection{Scale-dependent regularizer,  $R_{T{\mathbf{Y}}}$}
Consider the following optimization problem in the online setting:
\begin{align}\label{eq_online_TY}
\mathbf{y}_T \leftarrow \arg\min_{\substack{\mathbf{y}_T}} \left\| \mathbf{X}^\top\mathbf{X} - \mathbf{Y}^\top\mathbf{Y} - \alpha T \mathbf{I}_T \right\|_F^2.
\end{align}
By expanding the squared norm and keeping only the terms that depend on $y_T$, the problem is equivalent to:
\begin{align}
\mathbf{y}_T \leftarrow \arg\min_{\substack{\mathbf{y}_T}} & \phantom{+} [ -4\mathbf{x}_T^\top \left(\sum_{t=1}^{T-1} \mathbf{x}_t\mathbf{y}_t^\top\right) \mathbf{y}_T , \nonumber \\
& + 2\mathbf{y}_T^\top (\sum_{t=1}^{T-1}\mathbf{y}_t\mathbf{y}_t^\top+ \alpha T\mathbf{I}_m) \mathbf{y}_T, \nonumber \\
& - 2 \|\mathbf{x}_T\|^2 \|\mathbf{y}_T\|^2 + \|\mathbf{y}_T\|^4 ].
\end{align}
In the large-$T$ limit, the last two terms can be ignored since the first two terms of order $T$ dominates. The remaining cost function is a positive quadratic form in $\mathbf{y}_T$ and we could find the minimum by solving the system of linear equations:
\begin{align}\label{fp}
\left (\sum_{t=1}^{T-1} \mathbf{y}_t\mathbf{y}_t^\top + \alpha T \mathbf{I}_m \right) \mathbf{y}_T = \left( \sum_{t=1}^{T-1} \mathbf{y}_t\mathbf{x}_t^\top\right)\mathbf{x}_T.
\end{align}
Out of various ways to solve Eq. \eqref{fp}, we choose the weighted Jacobi iteration because it leads to an algorithm implementable by a biologically plausible neural network \cite{pehlevan2015hebbian}. In this algorithm,
\begin{equation}\label{eq_jacobi}
\mathbf{y}_T \leftarrow (1-\eta) \mathbf{y}_T + \eta (\mathbf{W}_T^{YX} \mathbf{x}_T - \mathbf{W}_T^{YY}\mathbf{y}_T),
\end{equation}
where $\eta$ is the weight parameter, and $\mathbf{W}_T^{YX}$ and $\mathbf{W}_T^{YY}$ are normalized input-output and output-output covariances:
\begin{align}\label{TYweights}
W_{T, ik}^{YX} &= \frac{\sum_{t=1}^{T-1} y_{t,i}x_{t,k}}{\alpha T + \sum_{t=1}^{T-1} y_{t,i}^2}, \nonumber \\
W_{T, ij\neq i}^{YY} &= \frac{\sum_{t=1}^{T-1} y_{t,i}y_{t,j}}{\alpha T + \sum_{t=1}^{T-1} y_{t,i}^2}, \nonumber \\
W_{T, ii}^{YY} &= 0.
\end{align}
When the Jacobi iteration converges to a fixed point, it obtains the solution of the quadratic program.

Such algorithm can be implemented by the dynamics of neural activity in a single-layer network. $\mathbf{W}_T^{YX}$ and $\mathbf{W}_T^{YY}$ represent the weights of feedforward $(\mathbf{x}_t \rightarrow \mathbf{y}_t)$ and lateral $(\mathbf{y}_t \rightarrow \mathbf{y}_t)$ synaptic connections. At each time step $T$, we first iterate~\eqref{eq_jacobi} until convergence, then update the weights online according to the following learning rules:
\begin{align}\label{eq_online_update_TY}
\mu_{T+1, i} &\leftarrow \mu_{T, i} + \alpha + y_{T,i}^2 \nonumber \\
W_{T+1,ij}^{YX} &\leftarrow W_{T,ij}^{YX} + \frac{(y_{T,i} x_{T,j} - (\alpha+y_{T,i}^2) W_{T,ij}^{YX})}{\mu_{T+1,i}} \nonumber \\
W_{T+1,ij\neq i}^{YY} &\leftarrow W_{T,ij}^{YY} + \frac{(y_{T,i} y_{T,j} - (\alpha+y_{T,i}^2) W_{T,ij}^{YY})}{\mu_{T+1,i}}.
\end{align}
where we introduce scalar variables $\mu_{T,i}$ representing cumulative activity of neuron $i$ up to time $T-1$. The left figure of Fig.~\ref{fig:network} illustrates this network implementation.
% fixed \todoyuansi{CP: 1) Use a different variable for the learning rate D, maybe $\eta$, because  we use D for the diagonal matrix. No need for ${}^Y$ superscript here.  2) we need a figure for this algorithm}.

\subsection{Input-output regularizer, $R_{\mathbf{XY}}$}

The online optimization problem with input-output regularizer is similar to the previous one. At every time step, the amount of thresholding is given by the cumulative sum of input eigenvalues. After slight modifications, we arrive at the following neural network algorithm. At time step $T$, $\mathbf{y}_T$ is iterated until convergence based on the following Jacobi iteration
\begin{align}\label{eq_jacobi_XY}
\mathbf{y}_T \leftarrow (1-\eta) \mathbf{y}_T + \eta (\mathbf{W}_T^{YX} \mathbf{x}_T - \mathbf{W}_T^{YY} \mathbf{y}_T).
\end{align}
The online synaptic updates are:
\begin{align}\label{eq_online_XY}
\mu_{T+1, i} &\leftarrow \mu_{T, i} + \alpha \|\mathbf{x}_T\|^2 + y_{T,i}^2 \nonumber \\
W_{T+1,ij}^{YX} &\leftarrow W_{T,ij}^{YX} + \frac{(y_{T,i} x_{T,j} - (\alpha \|\mathbf{x}_T\|^2+y_{T,i}^2) W_{T,ij}^{YX})}{\mu_{T+1,i}} \nonumber \\
W_{T+1,ij\neq i}^{YY} &\leftarrow W_{T,ij}^{YY} + \frac{(y_{T,i} y_{T,j} - (\alpha \|\mathbf{x}_T\|^2 +y_{T,i}^2) W_{T,ij}^{YY})}{\mu_{T+1,i}}.
\end{align}
This differs from Eq. \eqref{TYweights} in that a new scalar variable $\|\mathbf{x}_T\|^2$ is needed that sums up the current input amplitude across all input neurons. Biologically, in an Hebbian/anti-Hebbian neural network, such summation could be implemented via extracellular space or glia (Fig.~\ref{fig:network}, middle).

\subsection{Squared-output regularizer, $R_{\mathbf{YY}}$}\label{sec_online_YY}
Finally, we consider the following online optimization problem:
\begin{align}\label{eq_online_YY}
\mathbf{y}_T \leftarrow \arg\min_{\substack{\mathbf{y}_T}} \|\mathbf{X}^\top\mathbf{X} - \mathbf{Y}^\top\mathbf{Y} \|_F^2 + \alpha \text{Tr}(\mathbf{Y}^\top\mathbf{Y})^2.
\end{align}
In the large-T limit, $\sum_{t=1}^{T-1} \|\mathbf{y}_t\|^2 \gg \|\mathbf{y}_T\|^2$, we could instead solve a simplified version to the problem:
\begin{align}\label{eq_online_update_YY}
\mathbf{y}_T &\leftarrow \arg\min_{\substack{\mathbf{y}_T}} \|\mathbf{X}^\top\mathbf{X} - \mathbf{Y}^\top\mathbf{Y}\|_F^2 \nonumber \\
&\qquad\qquad+ 2 \alpha \text{Tr}(\mathbf{Y}_{T-1}^\top\mathbf{Y}_{T-1}) \text{Tr}(\mathbf{Y}^\top\mathbf{Y}).
\end{align}
After this simplification, the update is similar to the previous input-output regularizer except that the learning rate depends on the norm of current output vector, $\|\mathbf{y}_T\|^2$. Since this is a scalar, such summation can be easily implemented in biology using summation in extracellular space or glia (Fig.~\ref{fig:network}, right). At time step $T$ , $\mathbf{y}_T$ the neural network dynamics iterates until convergence of the Jacobi iteration~\eqref{eq_jacobi}. After convergence, synaptic weights are updated online as follows,
\begin{align}\label{eq_online_YY}
\mu_{T+1, i} &\leftarrow \mu_{T, i} + \alpha \|\mathbf{y}_T\|^2 + y_{T,i}^2 \nonumber \\
W_{T+1,ij}^{YX} &\leftarrow W_{T,ij}^{YX} + \frac{(y_{T,i} x_{T,j} - (\alpha \|\mathbf{y}_T\|^2+y_{T,i}^2) W_{T,ij}^{YX})}{\mu_{T+1,i}} \nonumber \\
W_{T+1,ij\neq i}^{YY} &\leftarrow W_{T,ij}^{YY} + \frac{(y_{T,i} y_{T,j} - (\alpha \|\mathbf{y}_T\|^2 +y_{T,i}^2) W_{T,ij}^{YY})}{\mu_{T+1,i}}.
\end{align}

 %%%%%%%%%%%%%%%%%%%%%%%%%%%%%%%%%%%%%%%%%%%%%%%%%%%%%%%%%%%%%%%%%%%%%%%%%%%%%%%%

% 5. Online adaptive dimension reduction for non-stationary statistics
\section{Online algorithms for non-stationary statistics}

The online algorithms we proposed assume that the input has stationary statistics. A truly adaptive algorithm, in addition to self-calibrating the number of dimensions to transmit, should be able to adapt to temporal statistics changes.  To address this issue, we introduce discounting into the cost function which reduces the contribution of older data samples \cite{pehlevan2015hebbian,yang1995projection}, or equivalently ``forgets" them, in order to react to changes in input statistics.

\subsection{Scale-dependent regularizer, $R_{T\mathbf{Y}}$}

Following \cite{pehlevan2015hebbian}, we discount past inputs, ${\bf x}_t$, and past outputs, ${\bf y}_t$, with $\beta^{T-t}$, where $0\leq\beta\leq 1$. With such discounting, the effective time scale of forgetting is $-1/\ln \beta$. This procedure leads to a modified online cost function,
\begin{align}\label{eq_online_forget_TY}
\mathbf{y}_T \leftarrow \arg\min_{\substack{\mathbf{y}_T}} \left\| B \mathbf{X}^\top\mathbf{X} B - B \mathbf{Y}^\top\mathbf{Y} B - \alpha \mathrm{Tr}(B^2) \right\|_F^2
\end{align}
where $B$ is a diagonal matrix with $(\beta^0, \beta^1, \beta^2, ..., \beta^{T-1})$ on the diagonal. The term $\mathrm{Tr}(B^2)$ takes the place of the time variable $T$  in the original online formulation \ref{eq_online_TY}. When $\beta$ is $1$, \eqref{eq_online_forget_TY} reduces to \eqref{eq_online_TY}.
% fixed \todoyuansi{CP: Add something about why Tr(B^2) is the right way to discount T.}

To derive an online algorithm, we follow the same steps as before. By keeping only the terms that depend on current output $\mathbf{y}_T$, we again arrive at a quadratic function of $\mathbf{y}_T$, which is solved as in \eqref{eq_jacobi} by a weighted Jacobi iteration. The online synaptic learning rules get modified:
\begin{align}\label{eq_rule_forget_TY}
\mu_{T+1, i} &\leftarrow \beta^2 \mu_{T, i} + \alpha + y_{T,i}^2, \nonumber \\
W_{T+1,ij}^{YX} &\leftarrow W_{T,ij}^{YX} + \frac{(y_{T,i} x_{T,j} - (\alpha+y_{T,i}^2) W_{T,ij}^{YX})}{\mu_{T+1,i}}, \nonumber \\
W_{T+1,ij\neq i}^{YY} &\leftarrow W_{T,ij}^{YY} + \frac{(y_{T,i} y_{T,j} - (\alpha+y_{T,i}^2) W_{T,ij}^{YY})}{\mu_{T+1,i}}.
\end{align}
The difference from the non-discounted learning rules (Eq. \eqref{TYweights}) is in how $\mu_{T, i}$ gets updated. The $\beta^2$ decay in $\mu_{T, i}$ update prevents $\mu_{T, i}$ from growing indefinitely. Consequently, the learning rate, $1/\mu_{T+1, i}$ does not steadily decrease with $T$, but saturates, allowing the synaptic weights to react to changes in input statistics.

\subsection{Input-output regularizer, $R_{\mathbf{XY}}$}
We can implement forgetting in this case again by discounting past inputs and outputs:
\begin{align}
\mathbf{y}_T \leftarrow \arg\min_{\substack{\mathbf{y}_T}} \left\| B \mathbf{X}^\top\mathbf{X} B - B \mathbf{Y}^\top\mathbf{Y} B - \alpha \mathrm{Tr}(B \mathbf{X}^\top \mathbf{X} B) \right\|_F^2.
\end{align}
Following the same steps as before, a weighted Jacobi iteration~\eqref{eq_jacobi} is still deployed. The following learning rules can be derived:
\begin{align}\label{eq_rule_forget_XY}
\mu_{T+1, i} &\leftarrow \beta^2 \mu_{T, i} + \alpha \|\mathbf{x}_T\|^2 + y_{T,i}^2, \nonumber \\
W_{T+1,ij}^{YX} &\leftarrow W_{T,ij}^{YX} + \frac{(y_{T,i} x_{T,j} - (\alpha \|\mathbf{x}_T\|^2+y_{T,i}^2) W_{T,ij}^{YX})}{\mu_{T+1,i}},  \nonumber \\
W_{T+1,ij\neq i}^{YY} &\leftarrow W_{T,ij}^{YY} + \frac{(y_{T,i} y_{T,j} - (\alpha \|\mathbf{x}_T\|^2 +y_{T,i}^2) W_{T,ij}^{YY})}{\mu_{T+1,i}}.
\end{align}
Again, $\beta^2$ decay in $\mu_{T,i}$ update allows the network to react to non-stationarity.

\subsection{Squared-output regularizer, $R_{\mathbf{YY}}$}
Discounting past inputs and outputs, we arrive at the online cost function:
\begin{align}
\mathbf{y}_T &\leftarrow \arg\min_{\substack{\mathbf{y}_T}} \left\| B \mathbf{X}^\top\mathbf{X} B - B \mathbf{Y}^\top\mathbf{Y} B\right\|_F^2 \nonumber \\
&\qquad\qquad + \alpha \mathrm{Tr}(B \mathbf{Y}^\top \mathbf{Y} B)^2
\end{align}
Following the same steps as before, a weighted Jacobi iteration~\eqref{eq_jacobi} is still deployed. The following learning rules can be derived:
\begin{align}\label{eq_rule_forget_YY}
\mu_{T+1, i} &\leftarrow \beta^2 \mu_{T, i} + \alpha \|\mathbf{y}_T\|^2 + y_{T,i}^2, \nonumber \\
W_{T+1,ij}^{YX} &\leftarrow W_{T,ij}^{YX} + \frac{(y_{T,i} x_{T,j} - (\alpha \|\mathbf{y}_T\|^2+y_{T,i}^2) W_{T,ij}^{YX})}{\mu_{T+1,i}},  \nonumber \\
W_{T+1,ij\neq i}^{YY} &\leftarrow W_{T,ij}^{YY} + \frac{(y_{T,i} y_{T,j} - (\alpha \|\mathbf{y}_T\|^2 +y_{T,i}^2) W_{T,ij}^{YY})}{\mu_{T+1,i}}.
\end{align}
%
%%%%%%%%%%%%%%%%%%%%%%%%%%%%%%%%%%%%%%%%%%%%%%%%%%%%%%%%%%%%%%%%%%%%%%%%%%%%%%%%

% Experiments
\section{Experiments}
We evaluate the performance of three online algorithms on a synthetic dataset. We first generate a $n=64$ dimensional colored Gaussian process with a specified covariance matrix. In this covariance matrix, the eigenvalues, $\lambda_{1..4} = \{6, 5, 4, 2\}$ and the remaining $\lambda_{5..60}$ are chose uniformly from the interval $[0, 0.2]$. Correlations are introduced in the covariance matrix by generating random orthonormal eigenvectors. We set $\eta$ to be $0.1$.  Synaptic weight matrices were initialized randomly.

Initially, we choose different $\alpha$'s for three algorithms such that the thresholding has the approximately the same effect on the original data: the output keeps track of the top three principal components, while discarding the rest principal components. Additionally, the top three eigenvalues of the input similarity matrix are soft-thresholded by $2$.
\subsection{Stationary input}
\begin{figure*}[ht]
\centering
\includegraphics[width=0.73\textwidth]{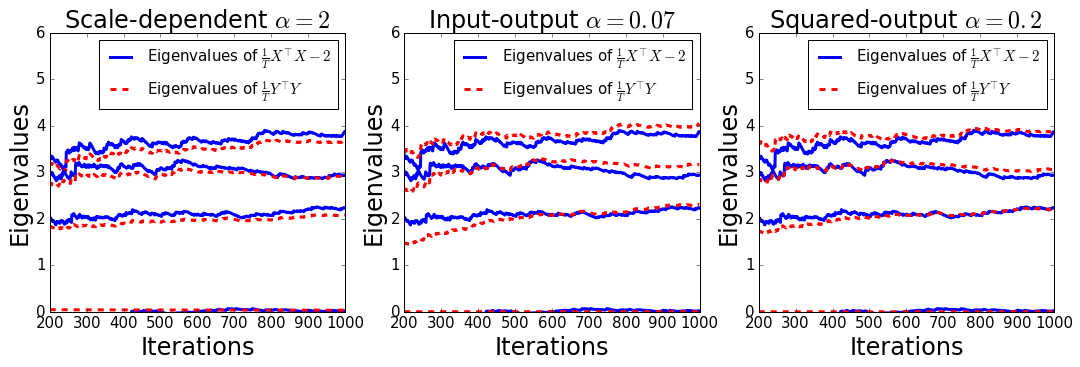}
\caption{All three online algorithms keep track of $\frac{1}{T} X^\top X$ top three principal components. Different $\alpha$'s are chosen such that top input eigenvalues are thresholded by $2$. From top to bottom, the four blue lines are corresponds to the top four eigenvalues of $\frac{1}{T} X^\top X - 2$. }
\label{fig:eigen_track}
\end{figure*}

As shown in Fig.\ref{fig:eigen_track}, with appropriately chosen $\alpha$, all three online algorithms are able to keep track of the top three input eigenvalues correctly.

\begin{figure}
    \centering
    \begin{minipage}{0.5\columnwidth}
        \includegraphics[width=1\textwidth]{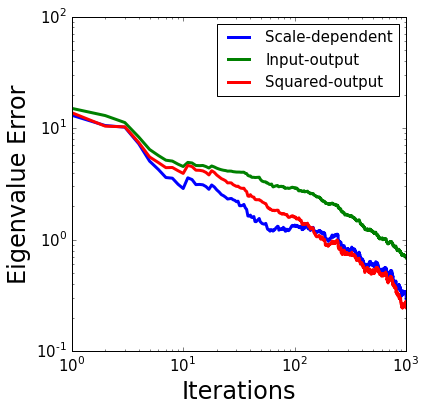}
        \label{fig:eigen_error}
    \end{minipage}%
    \begin{minipage}{0.5\columnwidth}
        \includegraphics[width=1\textwidth]{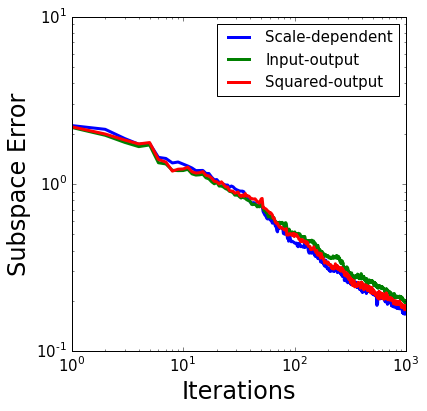}
        \label{fig:subspace_error}
    \end{minipage}
    \caption{Left: eigenvalue error as a function of iteration for three algorithms; Right: subspace error as a function of iteration for three algorithms.}
    \label{fig:eigen_subspace_error}
\end{figure}

To quantify the performance of these algorithms more precisely than looking at individual eigenvalues, we use two different metrics. The first metric, eigenvalue error, measures the deviation of output covariance eigenvalues from their optimal offline values derived in Section II. The eigenvalue error at iteration $T$ is calculated by summing squared differences between the eigenvalues of $\frac{1}{T} Y^\top Y$. The second metric, subspace error, quantifies the deviation of the learned subspace from the input principal subspace. The exact formula for the subspace error metric can be found in~\cite{pehlevan2015normative}. Fig~\ref{fig:eigen_subspace_error} shows that three algorithms perform similarly in terms of these two metrics. Both errors for each algorithm decrease as a function of iterations $T$.

\subsection{Non-stationary input}
We evaluate the performance of three online algorithms with forgetting on non-stationary input. The non-stationary input we use here has a sudden change of input statistics. We first use the original data generation process for $1000$ iterations. $\alpha$ is chosen for each algorithm such that the top three principal components are retained and the rest are discarded. Then we change the input data generation by multiplying the eigenvalues of $\mathbf{X}^\top\mathbf{X}$ by $2$, in order to see whether the algorithms can still keep track of only the top three principal components. Finally at $6000$ iteration, we change back to the original statistics. Since the input statistics changes over time, $\frac{1}{T}\mathbf{X}^\top\mathbf{X}$ is not reflective of the eigenvalues our online algorithms are tracking at time $T$. Thus we the eigenvalues over a short period $T_0$ of data before $T$.

\begin{figure*}
\centering
\includegraphics[width=0.73\textwidth]{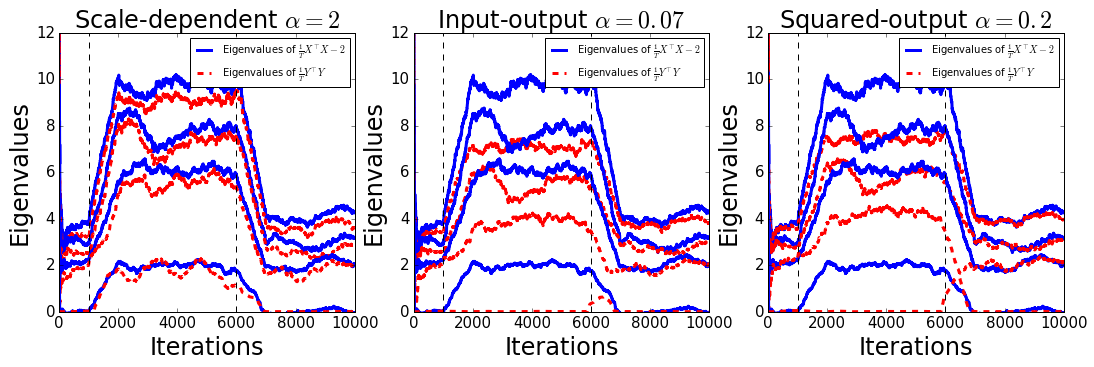}
\caption{Plot of eigenvalues of $\frac{1}{T_0}\mathbf{X}^\top\mathbf{X}$ and $\frac{1}{T_0}\mathbf{Y}^\top\mathbf{Y}$ over a short period $T_0 = 1000$ as a function of $T$. Eigenvalues of $\frac{1}{T}\mathbf{X}^\top\mathbf{X}$ are multiplied by $2$ at iteration $1000$. Then eigenvalues of $\frac{1}{T}\mathbf{X}^\top\mathbf{X}$ are scaled back at iteration $6000$. Scale-dependent regularizer has an output similarity matrix with four positive eigenvalues starting from iteration $1000$.}
\label{fig:sudden2}
\end{figure*}

For the first $1000$ iterations, all three online algorithms keep track of the top three principal components (See Fig. \ref{fig:sudden2}).  The fourth output singular value (fourth red line) is kept zero all the time, while the top three singular values (top three red lines) are above zero. At $1000$ iteration, there is sudden change of input data generation. The fourth output singular value for input-output regularizer and squared-output regularizer remains zero, however, the fourth output singular value for scale-dependent regularizer becomes larger than zero (See Fig. \ref{fig:sudden2}). Scale-dependent regularizer now has an output with effective dimension four rather than three. The other two are doing a better job in keeping track of only three principal components.

When the input data generation is changed back to the original one at iteration $6000$, because of the forgetting mechanism we introduced, all three regularizers are able to keep track of the top three principal components like during the first $1000$ iterations.

% \begin{figure}[ht]
% \includegraphics[width=\columnwidth]{figure/sudden1.png}
% \caption{Eigenvalues of $\frac{1}{T} \mathbf{X}^\top\mathbf{X}$ multiplied by $2$ at iteration $1000$. Scale-dependent regularizer has an output similarity matrix with four positive eigenvalues starting from iteration $1000$. }
% \label{fig:sudden1}
% \end{figure}

% \subsection{Numerical simulations: large $\alpha$: where the performances of $R_{\mathbf{XY}}$  and $R_{\mathbf{YY}}$ differ}

%%%%%%%%%%%%%%%%%%%%%%%%%%%%%%%%%%%%%%%%%%%%%%%%%%%%%%%%%%%%%%%%%%%%%%%%%%%%%%%%

% Conclusion
\section{Conclusion}

We have introduced online dimensionality reduction algortihms with self-calibrating regularizers. Unlike the scale-dependent adaptive dimensionality reduction algorithm \cite{pehlevan2015normative}, these self-calibrating algorithms are designed to automatically adjust to the variation in singular values of the input. As a consequence, they may be more appropriate for modeling neuronal circuits or any related artificial signal processing systems.

%%%%%%%%%%%%%%%%%%%%%%%%%%%%%%%%%%%%%%%%%%%%%%%%%%%%%%%%%%%%%%%%%%%%%%%%%%%%%%%%
\section*{APPENDIX} \label{Appendix}

\subsection{Proof: offline solution of squared-output regularizer} \label{Appendix:proof}

First, we cite a lemma from \cite{pehlevan2015normative}:

\noindent{\bfseries  Lemma 1}: Let $\mathbf{\Lambda} = \mathrm{diag}(\lambda_1,...,\lambda_p)$, where $\lambda_1 \geq ... \geq \lambda_p$ are real numbers, and let $\hat{\mathbf{\Lambda}} = \mathrm{diag}(\hat{\lambda}_1,...,\hat{\lambda}_p)$, where $\hat{\lambda}_1 \geq ... \geq \hat{\lambda}_p$ are real numbers. Then,
\[
\max_{\mathbf{O}\in O(p)} \mathrm{Tr}(\mathbf{\Lambda}\mathbf{O}\hat{\mathbf{\Lambda}}\mathbf{O}^\top) = \mathrm{Tr}(\mathbf{\Lambda}\hat{\mathbf{\Lambda}})
\]
where $O(p)$ is the set of $p\times p$ orthogonal matrices.

This lemma states that identity belongs to the optimal orthogonal transformations for diagonal matrix alignment. A complete proof of the lemma can be found in \cite{pehlevan2015normative}.

Offline adaptive soft-thresholding optimization problem has the following form:
\[
\min_{\mathbf{Y}} \left\| \mathbf{X}^\top\mathbf{X} - \mathbf{Y}^\top\mathbf{Y} \right\|_F^2 +  \alpha [\text{Tr}(\mathbf{Y}^\top\mathbf{Y})]^2.
\]
Suppose an eigen-decomposition of $\mathbf{Y}^\top\mathbf{Y}$ is $\mathbf{Y}^\top\mathbf{Y} = \mathbf{V}^Y\mathbf{\Lambda}^Y{\mathbf{V}^Y}^\top$. Since the Frobenius norm is invariant to multiplication of unitary matrices, we could multiply on left ${\mathbf{V}^X}^\top$ and on right ${\mathbf{V}^X}$ to obtain an equivalent objective
\[
\min_{\mathbf{\Lambda}^Y \geq 0, \mathbf{G}\in O(T)} \left\| \mathbf{\Lambda}^X - \mathbf{G}\mathbf{\Lambda}^Y\mathbf{G}^\top \right\|_F^2 +  \alpha [\text{Tr}(\mathbf{\Lambda}^Y)]^2.
\]
According to {\bfseries Lemma 1}, we conclude that $\mathbf{G}$ could be take $\mathbf{I}_T$ at optimum. Observing that $\text{Tr}(\mathbf{\Lambda}^Y)$ can be written as a linear transform of its diagonal elements. Then the remaining optimization on the diagonal matrix $\mathbf{\Lambda}^Y$ could be written as
\[
\min_{\mathbf{d}^Y \geq 0} \left\| \mathbf{d}^X - \mathbf{d}^Y \right\|^2 + \alpha (\mathbf{1}^\top\mathbf{d}^Y)^2.
\]
where $\mathbf{d}^X$ and $\mathbf{d}^Y$ are diagonals of $\mathbf{\Lambda}^X$ and $\mathbf{\Lambda}^Y$ respectively. We have an extra constraint that less than $n$ coordinates of $\mathbf{d}^Y$ could be nonzero. The problem could be written equivalently,
\[
\min_{\mathbf{d}_n^Y \geq 0} {\mathbf{d}_n^Y}^\top (\mathbf{I}_n + \alpha \mathbf{1}_n\mathbf{1}_n^T) \mathbf{d}_n^Y - 2 {\mathbf{d}_n^X}^\top \mathbf{d}_n^Y.
\]
where $\mathbf{d}_n^Y$ are the first $n$ elements of $\mathbf{d}^Y$.

This is a nonnegative least squares problem (NNLS), which has the general form
\[
\min_{x\geq 0} \left\| \mathbf{b} - \mathbf{A}\mathbf{x}\right\|^2.
\]
For general form of $\mathbf{A}$, this NNLS does not allow for closed form solution. In general, it is solved by an active-set type optimization algorithm~\cite{nocedal2006numerical}, and the number of iterations in the worse case could be exponential on the input dimension. In our case, $\mathbf{A} = \mathbf{I}_n + \alpha \mathbf{1}_n\mathbf{1}_n^\top$ and it almost allows for an closed form solution.

\begin{enumerate}
\item Since $\mathbf{A}$ is a diagonal matrix plus a constant matrix, when the values of $\mathbf{d}^X$ is ordered, the values of $\mathbf{d}^Y$ is also ordered.
\item Once the support of $\mathbf{d}^Y$ is known, the problem is a unconstrained positive definite quadratic program, which always allows for a closed form solution.
\item Combining (1) and (2), the support of the solution is always the first $p$ elements. It is sufficient to try $n$ different supports and find the best feasible solution.
\end{enumerate}

Now suppose that we have found that the support of solution is of size $p$, we could obtain a closed form solution of the offline problem. Given the support, the NNLS problem is equivalent to the unconstrained quadratic problem.
\[
\min_{\mathbf{d}_p^Y} {\mathbf{d}_p^Y}^\top (\mathbf{I}_p + \alpha \mathbf{1}_p\mathbf{1}_p^T) \mathbf{d}_p^Y - 2 {\mathbf{d}_p^X}^\top \mathbf{d}_p^Y.
\]
Solving the unconstrained quadratic problem, we obtain
\[
\mathbf{d}_p^Y = (\mathbf{I}_p - \frac{\alpha}{1+\alpha p} \mathbf{1}_p\mathbf{1}_p^\top) \mathbf{d}_p^X,
\]
since $\mathbf{A}$ is invertible with inverse $\mathbf{I}_p - \frac{\alpha}{1+\alpha p} \mathbf{1}_p\mathbf{1}_p^\top$.

\subsection{Performance of three regularizers in various signal and noise setups} \label{Appendix:performance}
%Here we illustrate that a universal regularization coefficient $\alpha$ will not always work for scale-dependent regularizer, while an appropriately chosen universal regularization $\alpha$ will almost work for the other two regularizers in any signal and noise setups.

%We compare the three regularized methods in the simple case of Section III. In addition, various signal and noise setups are used to check the robustness of a single universal regularization $\alpha$.
The set of signal and noise setups $(a, b)$ corresponds to the following $5050$ cases
\[
S = \{ (a, b) | a \geq b, (a, b) \in \{0.01, 0.02, ..., 0.99, 1.00\}^2\}.
\]

%\addtolength{\textheight}{-12cm}   % This command serves to balance the column lengths
                                  % on the last page of the document manually. It shortens
                                  % the textheight of the last page by a suitable amount.
                                  % This command does not take effect until the next page
                                  % so it should come on the page before the last. Make
                                  % sure that you do not shorten the textheight too much.

\section*{ACKNOWLEDGMENT}
We would like to thank Anirvan Sengupta for discussions.

% Generated by IEEEtran.bst, version: 1.13 (2008/09/30)

\end{document}